\documentclass{article}

\usepackage[margin=1in]{geometry} 
\usepackage{graphicx}
\usepackage{listings}
\usepackage{hyperref}
\usepackage{comment}

\usepackage{amsmath}
\usepackage{amsfonts}

\usepackage{tikz}
\usetikzlibrary{cd,arrows,calc,shapes,positioning}
\tikzstyle{obs} = [circle,fill=white,draw=black,inner sep=0pt,minimum size=18pt,font=\fontsize{10}{10}\selectfont,node distance=1,thick]

\tikzstyle{obsdummy} = [circle,fill opacity=0,draw=none,inner sep=0pt,minimum size=18pt,font=\fontsize{10}{10}\selectfont,node distance=1,thick]

\tikzstyle{hobs} = [circle,fill=white,draw=black,inner sep=0pt,minimum size=18pt,font=\fontsize{10}{10}\selectfont,node distance=1,ultra thick]
\tikzstyle{intv} = [rectangle,fill=white,draw=black,inner sep=0pt,minimum size=18pt,font=\fontsize{10}{10}\selectfont,node distance=1,thick]
\tikzstyle{hintv} = [rectangle,fill=white,draw=white,inner sep=0pt,minimum size=18pt,font=\fontsize{10}{10}\selectfont,node distance=1,ultra thick]
\tikzstyle{fact} = [rectangle,fill=black,draw=black,inner sep=0pt,minimum size=10pt,font=\fontsize{10}{10}\selectfont,node distance=1,thick]
\tikzstyle{latent} = [circle,fill=white,draw=black,inner sep=0pt,minimum size=18pt,font=\fontsize{10}{10}\selectfont,node distance=1,thick,dotted]

\newcommand{\edge}[3][]{ %
  \foreach \x in {#2} { %
    \foreach \y in {#3} { %
      \path (\x) edge [->, >={Latex[round]}, #1,thick] (\y) ;%
    } ;
  } ;
}

\DeclareMathOperator{\indep}{\perp\!\!\!\perp}
\DeclareMathOperator{\dep}{\not\! \perp\!\!\!\perp}

\title{Counterfactual Fairness Is Not Demographic Parity, \\and Other Observations}

\author{Ricardo Silva\\University College London\\ \tt{ricardo.silva@ucl.ac.uk} }
\date{}

\begin{document}

\maketitle

\abstract{
 Blanket statements of equivalence between causal concepts and purely probabilistic concepts should be approached with care. In this short note, I examine a recent claim that counterfactual fairness is equivalent to demographic parity. The claim fails to hold up upon closer examination. I will take the opportunity to  address some broader misunderstandings about counterfactual fairness.
}

\section{Introduction}

\emph{This manuscript is motivated by \cite{rosenblatt:23}, published at AAAI 2023. It is based on some misunderstandings about counterfactual fairness, leading to the incorrect conclusion that counterfactual fairness and demographic parity are equivalent. Emphatically, the goal of this manuscript is \emph{not} to criticize any particular paper. Instead, it follows from the fact that none of the AAAI reviewers were able to help the authors of \cite{rosenblatt:23}. This suggests to me that will would be helpful to provide some notes on possible common missteps on understanding counterfactual fairness, using \cite{rosenblatt:23} just as a springboard for broader comments.}

\emph{In a nutshell, counterfactual fairness is a notion of individual fairness that lies on Rung 3 of Pearl’s ladder of causality \cite{pearl:18}, while demographic parity is a non-causal notion of group fairness occurring at the purely probabilistic Rung 1. We illustrate that there are scenarios of strong causal assumptions where the two notions ``coincide'', in an inconsequential equivalence that relies on narrow Rung 3 conditions. More generally, the equivalence breaks down. A simple way of seeing this is as follows: we already know that different causal models can induce different feasible regions of counterfactually fair predictors that do not coincide  despite being observationally and interventionally equivalent \cite{russell:17}. Hence, for any given predictor $\hat Y$ respecting demographic parity, an adversary can pick a particular world that explains the data but in which $\hat Y$ is not counterfactually fair. In this manuscript, we give constructive examples on how trivial it is to break the counterfactual fairness of any predictor which respects demographic parity but disregards counterfactual assumptions, among other clarifications of common misunderstandings. We then proceed to look closer at other problematic claims about counterfactual fairness described by \cite{fawkes:22} and \cite{plecko:22} that fall within the broader theme of how counterfactual fairness is sometimes misapplied.}

\emph{In the interest of keeping this note short, I will not cover any background material and definitions that can be commonly found in the literature. See \cite{barocas:23} for suitable background information.}

\section{On Counterfactual Fairness and Demographic Parity}
\label{sec:background}

\paragraph{Links between counterfactual fairness and demographic parity have been known from the beginning.} \noindent Relations between counterfactual fairness and demographic parity are known since the original paper  \cite{kusner:17}, which has an appendix section with the relevant title ``Relation to Demographic Parity'' (Appendix S2, page 13). That section was primarily concerned about a less standard notion of demographic parity -- the writing most definitely should have been clearer, as it muddles the presentation of the new variant with the more standard definition of demographic parity. The variant of demographic parity in that appendix was implicitly defined in terms of a predictor $\hat Y$ being functionally independent of protected attributes $A$, rather than the more standard (and non-causal) definition $\hat Y \indep A$. The latter can be  checked using textbook methods from e.g. the graphical modeling literature.

\paragraph{Counterfactual fairness does not imply demographic parity.} The focus on Appendix S2 of \cite{kusner:17} was on a functional variant of demographic parity, as I judged that interested readers can always deduce by themselves when  standard demographic parity ($\hat Y \indep A$) holds. Here's an illustration.

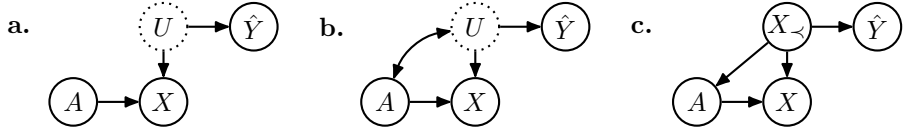
\begin{figure}[t]
\centering
  \begin{tabular}{ccc}
  \begin{tikzpicture}
    \node[right] (L) at (-1, 2) {\textbf{a.}};
    \node[obs] (A) at (0, 1) {$A$};
    \node[obs] (X) at (1.2, 1) {$X$};
    \node[latent] (U) at (1.2, 2) {$U$};
    \node[obs] (Yhat) at (2.4, 2) {$\hat Y$};
    \edge[black]{A}{X};
    \edge[black]{U}{X};
    \edge[black]{U}{Yhat};
  \end{tikzpicture}%
  &
  \begin{tikzpicture}
    \node[right] (L) at (-1, 2) {\textbf{b.}};
    \node[obs] (A) at (0, 1) {$A$};
    \node[obs] (X) at (1.2, 1) {$X$};
    \node[latent] (U) at (1.2, 2) {$U$};
    \node[obs] (Yhat) at (2.4, 2) {$\hat Y$};
    \edge[black]{A}{X};
    \edge[black]{U}{X};
    \edge[black]{U}{Yhat};
    \edge[<->, bend right=-30]{A}{U};
  \end{tikzpicture}%
  &
  \begin{tikzpicture}
    \node[right] (L) at (-1, 2) {\textbf{c.}};
    \node[obs] (A) at (0, 1) {$A$};
    \node[obs] (X) at (1.2, 1) {$X$};
    \node[obs] (X_p) at (1.2, 2) {$X_\prec$};
    \node[obs] (Yhat) at (2.4, 2) {$\hat Y$};
    \edge[black]{A}{X};
    \edge[black]{X_p}{X};
    \edge[black]{X_p}{Yhat};
    \edge[black]{X_p}{A};
  \end{tikzpicture}%
  \hspace*{1.0cm}
 \end{tabular}
  \caption{Three different acyclic directed mixed graphs encoding causal assumptions among protected attributes $A$, post-treatment variables $X$, pre-treatment variables $X_\prec$, and (latent) structural errors $U$. Variables ommitted where unnecessary/unavailable. The causal structures are augmented with the factor relating which variables are used in some predictor $\hat Y$, so that the Markovian structure of the system and predictor can be jointly read-off. Only (a) structurally implies demographic parity. The converse is not true even when $A$ is ignorable: an arbitrary predictor $\hat Y$ that satisfies demographic parity will in general not imply counterfactual fairness, as explained in the text.}
 \label{fig:dp}
\end{figure}

Figure \ref{fig:dp}(a) depicts a simple causal diagram where $A$ are protected attributes, $X$ are other measurements of an individual, $U$ is the error term for the structural equation of $X$, and $\hat Y$ is a predictor. Adding $\hat Y$ to the diagram is fine regardless whether any intervention on $U$ is well-defined, as the graph also has the semantics of a probabilistic independence model. This is the type of diagram that an experienced reader would be able to mentally construct by eyeballing Figure 1 of \cite{kusner:17}, for instance. In it, $\hat Y$ is d-separated from $A$, and hence $\hat Y \indep A$ for any model Markov with respect to that graph. In Figure \ref{fig:dp}(b), that's not the case, and demographic parity is not entailed by the structural assumptions. The same with Figure \ref{fig:dp}(c), where $X_\prec$ are pre-treatment variables, and so on. Key extensions to the bare-bones counterfactual fairness idea, such as path-specific versions that were alluded in the original paper and developed in detail in contributions such as \cite{chiappa:19}, also have no trivial connection to demographic parity.

What if we consider only inputs as in Figure \ref{fig:dp}(a), and only the basic total-effect counterfactual fairness criterion of the original paper? Let's call this setup ``special counterfactual fairness'' (sCP) to clarify we are dealing with a specific case. A key question is: if a predictor $\hat Y$ satisfies demographic parity and we happen to know that $A$ is ignorable (that is, $A \indep \{X_a\}$, where $\{X_a\}$ is the set of potential outcomes of $X$), does $\hat Y$ automatically satisfy counterfactual fairness? The answer is no.

\paragraph{Demographic parity does not imply counterfactual fairness.} Let's provide a simple counterexample where $X$ given $A$ is Gaussian and $A$ is 0/1 binary. These distributions are unimportant but will make the presentation easier. We will use $V_w$ to denote the potential outcome of variable $V$ under an intervention that sets some variable $W$ to a fixed level $w$. Any structural causal model (SCM) implies a joint distribution over potential outcomes \cite{pearl:09}.

Let $X_0$ and $X_1$ be the corresponding potential outcomes for $X$ under interventions on $A$, and assume that they are jointly Gaussian with means $\mu_0, \mu_1$, variances $\sigma^2_0, \sigma^2_1$, and correlation $\rho$. In SCM lingo, we could obtain that if the structural equation is e.g. $X = I(A = 0)(\mu_0 + U_0)+ I(A = 1) (\mu_1 + U_1)$, where $U_0, U_1$ are zero-mean Gaussian variables with the same covariance structure as $X_0, X_1$.

Let's define $\hat Y$ deterministically as the following function of $A$ and $X$, $$\displaystyle \hat Y := \frac{X - \mu_A}{\sigma_A}.$$ We can do this if we know (learn) the density $p(x~|~a)$, since the marginals of the potential outcomes are identifiable here\footnote{Yes, in principle we should define $\hat Y$ as a function $g((X - \mu_A) / \sigma_A)$ where $g$ minimizes some loss function with respect to some $p(y~|~a, x)$, but this is not relevant to the argument.}. It is clear that $\hat Y \indep A$, while $\hat Y \dep A~|~X$ and $\hat Y \dep X~|~A$ -- that's non-trivial demographic parity taking place. This also implies that $\hat Y_0$ and $\hat Y_1$ are equal in distribution, and that there is a one-to-one mapping between $\hat Y$ and $U_0$ when $A = 0$, and between $\hat Y$ and $U_1$ when $A = 1$. 

What about counterfactual fairness? The only remaining unidentifiable parameter is $\rho$. It would be \emph{terrific} if we were able to ensure counterfactual fairness with only the relatively mild assumption of ignorability of $A$. Alas, this is not meant to be:
\[
\begin{array}{rcl}
\displaystyle
P(\hat Y_1 \leq y~|~X = x, A = 0) &=&
P(X_1 \leq \sigma_1y + \mu_1~|~X = x, A = 0, X_0 = x)\\
\\
&=&
P(X_1 \leq \sigma_1y + \mu_1~|~X_0 = x)\\
\\
&\propto&\displaystyle
\int_{-\infty}^{\sigma_1y + \mu_1}
\exp\left\{-\frac{1}{2} \frac{(x^\star - f_1(x, \mu_0, \sigma_0, \mu_1, \sigma_1, \rho))^2}{f_2(\mu_0, \sigma_0, \mu_1, \sigma_1, \rho)} \right\}\, dx^\star,
\end{array}
\]
\noindent where $f_1(\cdot)$ and $f_2(\cdot)$ are non-trivial functions of $\rho$. The second line come from the consistency implication $\{A = 0, X_0 = x\} \Rightarrow X = x$ and the fact that by ignorability we have that $A \indep \{X_0, X_1\}$. 

This means that the above is in general different from
\[
\begin{array}{rcl}
\displaystyle
P(\hat Y_0 \leq y~|~X = x, A = 0) &=&
P(\hat Y \leq y~|~X = x, A = 0)\\
\\
&=& \displaystyle 1\left(\frac{x - \mu_0}{\sigma_0} \leq y\right),\\
\end{array}
\]
which, for starters, is a point-mass distribution. If for whatever reason we had chosen $\hat Y$ not to be a deterministic function of $(A, X)$, the factual marginal is still \emph{not} a function of $\rho$. For instance, if $\mu_0 = \mu_1 = 1$ and $\sigma^2_0 = \sigma^2_1 = 1$, then $f_1(x, \mu_0, \sigma_0, \mu_1, \sigma_1, \rho) = \rho \times x$ and $f_2(\mu_0, \sigma_0, \mu_1, \sigma_1, \rho) = 1 - \rho^2$, with the resulting integral for $P(\hat Y_1 \leq y~|~X = x, A = 0)$ evidently different from the binary function $1\left(x \leq y\right)$, and in general a function of $\rho$, even if $\hat Y$ is not a deterministic function of $(A, X)$.

This means counterfactual fairness is in general not implied by demographic parity even in the most basic sCF scenario.

The assumptions of Gaussianity are not fundamental. For a scalar continuous $X$, we know that we can parameterize any density $p(x~|~a)$ as $X = F_{x|a}^{-1}(U)$, where $U$ in uniformly distributed in $[0, 1]$ and $F_{x|a}^{-1}(\cdot)$ is the inverse of the conditional cdf $F_a(x) := P(X \leq x~|~A = a)$ (see \cite{rosenblatt:52} for a multivariate version of this idea). We can define (the distribution of) $\hat Y$ via $p(\hat y~|~u)$ for any (point-mass or not) conditional pmf/pdf of our choice, and thus we have that $\hat Y \indep A$ (the Markov blanket of $Y$ still includes both $A$ and $X$ when we marginalize $U$). This all follows directly from a Markov structure equivalent to the one in Figure \ref{fig:dp}(a). However, one big point is that \emph{here $U$ is an arbitrary mathematical construct and by itself warrants no interpretation as the error term of a structural causal model.}.

\paragraph{Remarks.} Even an expert in SCMs may get confused about the meaning of a structural equation such as $X = A + \epsilon$. The claim in this equation is that the error term is one-dimensional and shared by all potential outcomes. The set of potential outcomes $\{X_a\}$ forms a stochastic process, potentially infinite dimensional if $A$ is continuous, so something is being thrown away here\footnote{A one-dimensional error term can still represent a linear function of infinitely many variables -- say, the shamelessly vague ``sum of all other causes of $X$'' -- but it clearly won't imply an infinite-dimensional model. As an analogy, think of the Gaussian distribution being parametric even when motivated as the average of infinitely many random variables.}. 

Under $X = A + \epsilon$ all potential outcomes are deterministically related, as $X_a = a + \epsilon$ and therefore $X_{a'} = X_a - a + a'$. Unlike the example in the previous paragraphs, this model has no free parameters for the correlation of the potential outcomes. What is the implication?

Say the world behaves according to a causal independence model given by the causal DAG $A \rightarrow X$ (again, no confounding) with structural equation $X = A + \epsilon$, where $\epsilon$ is Gaussian, and so is $A$. Therefore, pdf $p(a, x)$ is the pdf of a bivariate Gaussian. Say also that an analyst knows $p(a, x)$, which is in principle learnable from observational data. 

The analyst sets themselves up to construct $\hat Y$ as a linear function of $(A, X)$ (and so also Gaussian) while enforcing $\hat Y \indep A$, using no more than $p(a, x)$, causality seemingly playing no role. This is akin to creating an unfaithful distribution \cite{sgs:00}, with the total effect of $A$ on $\hat Y$ being canceled by combining the path $A \rightarrow X \rightarrow \hat Y$ against $A \rightarrow \hat Y$. Let's assume $A$ and $X$ are marginally standard Gaussians for simplicity. From free parameters $(\lambda_1, \lambda_2, \lambda_3)$ in $\hat Y = \lambda_1 A + \lambda_2 X + \lambda_3$ and the covariance matrix of $(A, X)$, we can pick $\hat Y = X - A$ (or any linear function of $X - A$) in order to guarantee $\hat Y \indep A$. But that's just (a linear function of) $\epsilon$, and therefore it seems that from demographic parity only we were able to achieve counterfactual fairness.

Or were we? In fact, we were not. We are not showing anything like $$\text{\emph{Demographic Parity}} \Rightarrow \text{\emph{Counterfactual Fairness}}.$$ 

Instead, we are showing (a special case of) the far less interesting result $$\text{\emph{Strong Causal Assumptions}} \Rightarrow (\text{\emph{Demographic Parity}} \Rightarrow \text{\emph{Counterfactual Fairness}}).$$

This has a closer link to demographic parity, even if of dubious value added, than the initial set of assumptions about $A$ being ignorable and nothing else (which led to nowhere). However, even the link we just found would fail if the structural equation was instead $X = A + \epsilon_A$, where $\epsilon_A$ means that, after fixing $A$ to $a$, we pick $\epsilon_a$ out of a zero-mean infinite-dimensional Gaussian process indexed by the action space of $A$ (the real line). A discussion of finite-dimensional stochastic processes for counterfactuals with continuous-valued treatments is provided by \cite{padh:23}.

The difference between $X = A + \epsilon$ and $X = A + \epsilon_A$ with a stationary Gaussian process $\{\epsilon_a\}$ is undetectable by experiment, even if the choice matters for counterfactual fairness. We are really stuck in what Pearl calls the ``Rung 3'' of causal inference \cite{pearl:18}. Any claim to reduce it to ``Rung 1''-level (purely probabilistic) should be taken apart to find out where the Rung 3 assumptions are hiding. It would be awesome and a relief to be able to stick to Rung 1, but I cannot  even imagine a counterfactual world where that would be true. I do not doubt that there are other systematic families of counterfactual assumptions that could lead to e.g. demographic parity as an alternative construction rule for counterfactually fair predictors, although it is not clear in which practical sense this would make our life any easier than a method for extracting functions of error terms or potential outcomes. Such constructions could also end up being of no use in more general scenarios such as path-specific effects.

A closing comment in this section. The original paper \cite{kusner:17}, and some of the follow-ups, exploited one-dimensional error term models as an illustration of the ideal pipeline, without committing to it as a clearly universal counterfactual family. Before spurious statements such as ``one-dimensional error terms are among the most common found in counterfactual fairness papers, therefore we will take it as part of the definition of counterfactual fairness etc.'' start to show up as a follow-up to this note, let me quote directly from \cite{kusner:17}:

\begin{quote}
Although this is well understood, it is worthwhile remembering that causal models always require strong assumptions, even more so when making counterfactual claims [8]. Counterfactuals assumptions such as structural equations are in general unfalsifiable even if interventional data for all variables is available. This is because there are infinitely many structural equations compatible with the same observable distribution \dots, be it observational or interventional. Having passed testable implications, the remaining components of a counterfactual model should be understood as conjectures formulated according to the best of our knowledge. Such models should be deemed provisional and prone to modifications if, for example, new data containing measurement of variables previously hidden contradict the current model.
\end{quote}

Reference [8] in the quote above is a 20-year old paper by my colleague A.P. Dawid \cite{dawid:00}, a most merciless critique of the unnecessary use of counterfactuals/potential outcomes in causal models that in reality do not \emph{need} them. Dawid calls counterfactual assumptions downright metaphysical, as there are no possible experiments to falsify some of their most general implications. I think it is important to have this critique in mind\footnote{Like Philip Dawid -- who, post-retirement and pre-pandemic, still used to show up regularly at the visitor's office in our department at UCL, a couple of doors down the corridor from mine -- I do share a distaste for unnecessary usage of counterfactuals/potential outcomes. To be honest, I was never really bothered by their use when the point was merely to introduce convenient notation -- for instance, I'm fond of using them in partial identification methods even for no-cross-world (``single-world'') questions. But once upon a time there was a genre, thankfully in decline, of ill-informed papers (primarily in statistics) dismissing the soundness of causal graphical models on the account they did not imply the existence of potential outcomes. My bafflement was due to the fact that in many of these papers there was indeed no need for potential outcomes at all. I would privately joke that those papers should get a \emph{Doubly-William Award}, named after Occam and Shakespeare, for demanding counterfactuals in a context where they would boil down to no more than ``entities multiplied without necessity, full of sound and fury, signifying nothing.''}, but even Dawid agrees that there \emph{are} genuine uses where counterfactual models play a central role despite untestable assumptions about the mere existence (!) of counterfactuals. With the rise of conformal prediction \cite{vovk:05, angelopoulos:23} in particular, it is becoming more acceptable to complement point predictions with prediction intervals, and partially-identifiable counterfactual intervals (on top of aleatoric + estimation error intervals) are becoming more mainstream \cite{lei:21}. A direction worthwhile pursuing is on partial identifiability \cite{padh:23}, where the companion paper \cite{russell:17} was an early but primitive example, followed by far more interesting work such as \cite{wu:19}. More of that would be welcome.

\section{Comments on Rosenblatt and Witter (2023)}
\label{sec:rw}

\cite{rosenblatt:23} attempts to show equivalence between counterfactual fairness and demographic parity. One direction is clearly correct, where sCP implies demographic parity, as in Figure \ref{fig:dp}(a) (although the paper imprecisely equates sCP with counterfactual fairness and there are no comments about path-specific variants). Their Theorem 4 states the converse, that demographic parity implies counterfactual fairness. As we have seen, this is not correct\footnote{It's not always easy to understand what the exact claim is: sCP and demographic parity are meant to be ``basically'' equivalent in some unspecified sense of what ``basically'' means. In one passage it is stated that \emph{``Note that Proposition 9, combined with the fact that Listing 1 and 2 satisfy demographic parity and preserve group order, does not refute Theorem 4 since the method of estimating latent variables in the proof of the theorem uses a trivial causal model.''} To the best of my understanding, this (an example of a predictor respecting demographic parity but not sCP) seems to directly refute Theorem 4. The argument in Theorem 4 literally states that $\hat Y = \hat Y'$, so that \emph{any} predictor respecting demographic parity must be counterfactually fair under sCP, regardless or not whether the person designing $\hat Y'$ carries out in their brain the construction of an imaginary latent variable model and predictor $\hat Y$.}. The bug in their proof is the conflation of mathematical constructs with latent variables that have causal semantics, a type of reification, along with tautological reasoning. Their argument relies on ``estimating the latent variables'' (presumably this can include the potentially infinite dimensional error term for the structural equation of $X$) using some preliminary $\hat Y'$, then using this ``$\hat U$'' to define another $\hat Y$. This says nothing about conditional cross-world dependencies, and as a matter of fact it says nothing even about a correctly specified model $X = f(A, U)$ for which $\hat Y'$ could in any sense qualify as an ``estimate'' of $U$. Even if we are not talking about error terms, this is entirely contrary to, for example, the advice in \cite{kusner:17} when considering Level 2 modeling for counterfactual fairness (emphasis added),
\begin{quote}
    Postulate background latent variables that act as non-deterministic causes of observable variables, based on \emph{explicit domain knowledge} and learning algorithms.
\end{quote}

This is not to pick on the authors: reification errors are a very common mistake in the literature that many have made at some point, present company included.

Interestingly, \cite{rosenblatt:23} cites \cite{fawkes:22} as showing links to demographic parity. But the main point of \cite{fawkes:22} are the challenges of constructing causal models that correctly capture ignorability conditions, and in fact demographic parity is only mentioned as a \emph{counterexample} to Theorem 4, where demographic parity holds but counterfactual fairness does \emph{not} (due to model misspecification). We will briefly discuss some aspects of \cite{fawkes:22} in the next section.

There are a few other issues I could comment on\footnote{An old favorite zombie is the following argument: i) race (for example) is a complex concept made of many constitutive aspects; ii) the existence of an overall intervention controlling all of those aspects is not self-evident; iii) we therefore conclude that we cannot possibly talk about any meaningful aspect of race as a cause. As there is so much in the literature about this, and this is not about the dependency structure of $(A, X, \hat Y)$ which is the topic of this manuscript, I'll refrain from further comments. I'll just leave here the acerbic comment of \cite{shalizi:05}, footnote 2, Chapter 21: ``It may be merely coincidence that ... many of the statisticians who make such pronouncements [there is no meaning of race or sex as a cause] work or have worked for the Educational Testing Service, an organization with an interest in asserting that, strictly speaking, sex and race cannot have any causal role in the score anyone gets on the SAT.'' We should do better than indulging Merchants of Doubt. See \cite{zoh:23} for an example of how to properly lead a thoughtful discussion on the challenges of defining causal effects of composite variables.}, but in the interest of focus, I will keep to the topic of dependency structure of $(A, X, \hat Y)$ and basic ideas that follow from it that are sometimes overlooked.

\paragraph{Counterfactual fairness as a variable selection procedure.} One of the main motivations for counterfactual fairness is as follows. In a prediction problem, we do not want fairness through unawareness \cite{barocas:23}, that is, just to remove the protected attributes from being inputs to a predictive system. This is because other variables carry implications from protected attributes, where ``implications'' is a loaded word that carries multiple meanings. Value judgements are necessary: they are part of the input in a machine learning pipeline, to be communicated in a way it translates into \emph{removing} sensitive information from a predictor that is indirectly implied by the protected attributes. The operational meaning of ``removing'' is also not straightforward. Section 6 of \cite{hardt:16} does a good job of describing the challenges. In counterfactual fairness and its path-specific variants, this judgement is carried over by a notion of counterfactual effect, essentially operationalizing what needs to be removed and how. Regarding this goal, nothing could be more different than fairness through unawareness. It is puzzling that \cite{rosenblatt:23} uses fairness through unawareness as a substitute to the original Level 1 of \cite{kusner:17}, going as far as claiming that it ``come(s) from the original counterfactual fairness paper'', as the approaches could not be more diametrically opposite to each other. I also do not understand what Level 3 in \cite{rosenblatt:23} is doing, but Level 3 in \cite{kusner:17} is most definitely not meant to be a ``compromise'' between 1 and 2, as ambiguously stated. Statements such as ``Level 3 is an example of an algorithm that ostensibly satisfies counterfactual fairness'' (but fails to achieve demographic parity) are  meaningless if the algorithm is based on misspecified causal models -- in \cite{kusner:17} we explicitly indicate that experiments on fairness measures were based on simulated data (generated from a model fitted to real data).  The text in the experimental section of \cite{rosenblatt:23} seems not to distinguish a definition from the improper application of this definition to models that may be misspecified.

Counterfactual fairness defines \emph{constraints} on a predictor, not the loss function, constraints which can be interpreted as \emph{filters} or a variation on the idea of information bottleneck \cite{tishby:99}. In fact, we can build \emph{optimal} loss minimizers \emph{with respect to} the information that counterfactual fairness allows to be used, which brings us to our next point.

\paragraph{Counterfactual fairness is agnostic to loss functions and other constraints.} If one wants to build a counterfactually fair predictor by flipping a coin, they can. The loss will be high, but this does not contradict counterfactual fairness\footnote{It also does not contradict our previous statement about needing causal assumptions: that $A$ is unconfounded with coin flip at a Rung 3 level is a causal assumption that e.g. the physical concept of superdeterminism does not need to assume.}. The point is that counterfactual fairness \emph{allows} for the use of information in $X$ that other (causal) concepts do not, and vice-versa (for instance, \cite{nabi:18} does not allow for mediators to be used in any sense, while counterfactual fairness allows for mediators to be used, as long their role is in the inference of potential outcomes/error terms).

Once predictions are generated, \emph{decisions} can incorporate other elements e.g. for college admissions we may want to avoid extremely uneven distribution of students according to (say) a notion of non-ignorable social economical status (as triggered by some moral judgment that no automated method should pretend to address), and so we may stratify our student ranking based not only on graduation success predictions but other desiderata such as the minimum intake of students of a particular background that the institution is targeting. Sometimes the term ``decision'' and the term ``prediction'' are informally used interchangeably (see for instance our own abstract of \cite{kusner:17}). On a closer look, this creates confusion. In a broader sense, \emph{decisions and predictions are not the same thing}. This is why, when speaking of actionable decisions, we go beyond the usual counterfactual fairness definitions see e.g. \cite{kusner:19} and \cite{gultchin:24}. We can \emph{predict} a first-year average for a prospective applicant, predictions which are reasonably independent across units. We can then \emph{decide} how to allocate offers based partially on those predictions using a fixed rule that is not \emph{learned} from data, where decisions entangle the units as resource allocation bottlenecks will take place -- notice that the original counterfactual fairness, by design, is not meant to guide predictions that take into account information from multiple units. Yes, we \emph{could} design a prediction task whether the label is a ``decision'' (offer/no offer) based on historical decisions -- sometimes bizarrely justified as following the principles of empirical risk minimization (as if we knew which false negatives, i.e. ``wrong'' declines, took place!). But if you believe that historical decisions are biased, and that they are not independently distributed, then maybe, just maybe, you should consider that it is not a good idea to try to emulate them even if the goal is building a ``fair projection'' of a flawed decision stream (itself possibly generated by a spectrum of many mutually contradicting decision makers selected by no well-defined distribution). 

Of course, just because counterfactual fairness is agnostic to the use of other constraints, it is still of interest to know when their intersection is empty.

In particular, \cite{rosenblatt:23} introduces a potentially interesting concept named group ordering preservation, described as ``...relative ordering from a (possibly) unfair predictor should induce the same relative ordering as the fair predictor.'' This deserves further consideration, but as currently written, it remains partially unclear to me what it means. Two algorithms\footnote{The difference between Listing 1 and Listing 2 is that the former assumes conditionally Gaussian cdfs. In what follows, we will focus on the latter.}, ``Listing 1'' and ``Listing 2'', provide an implementation of this concept. In a nutshell, they take preliminary (``unfair'') labels $\bar y^{(i)}$ for each data point $i$ in the sample and return ``fair'' labels $\hat y^{(i)}$ defined by mapping the quantile of $\bar y^{(i)}$ in the cdf $F_{a^{(i)}}(\bar y^{(i)}) := P(\bar Y \leq \bar y^{(i)}~|~A = a^{(i)})$ to the same quantile in the marginal cdf $F(\bar y^{(i)}) := P(\bar Y \leq \bar y^{(i)})$. The idea is that, within any protected attributes stratum, ``fair labels'' should be ranked similarly to ``unfair labels'' once the unfair components are removed in some desirable sense (demographic parity, for the cases of Listing 1 and 2). 

To summarize my understanding, this means that we want $\hat Y$ to satisfy some fairness constraint that $\bar Y$ does not necessarily follow, while  enforcing that $\hat Y$ and $\bar Y$ have rank (Spearman) correlation of 1 within each stratum of $A$. In particular, the template for Listing 2-like algorithms starts from a preliminary ``unfair'' predictor $\bar Y^{(i)} = \bar y^{(i)}$ and produces
\begin{equation}
\hat y^{(i)} := g(F^{-1}(F_{a^{(i)}}(\bar y^{(i)}))),
\label{eq:real_deal}
\end{equation}
where $g$ is a increasingly monotonic function (say, the identity function). Notice that if $\bar Y$ respects both demographic parity and counterfactual fairness, then in the limit of infinite training data so will $\hat Y$. 

Proposition 9 of \cite{rosenblatt:23} attempts to claim that counterfactual fairness cannot in general obey the ``group ordering preservation'' criterion. It is not clear to me what this means, but the ``counterexample'' provided as the proof in Proposition 9 of \cite{rosenblatt:23} involves dependencies between units, and that is not even considered by counterfactual fairness. Presumably, there $\bar Y$ are the true labels (no other predictors are mentioned) and $\hat Y$ is a counterfactually fair predictor. It is of course impossible for any pair $(\hat Y, Y)$, regardless whether $\hat Y$ is counterfactually fair or not, to have a rank correlation of 1 if $Y$ is not a deterministic function of $(A, X)$ (the only information we can use in a prediction $\hat Y$) so it is not clear what the goal is. This also does not clarify why we would want to align our $\hat Y$ with some $\bar Y$ other than $Y$, different alignments which can be vastly incompatible. More about that is discussed in the appendix to this manuscript, where I present what I believe are fundamental issues with the main experiment in \cite{rosenblatt:23}.

\section{Comments on Fawkes, Evans and Sejdinovic (2022)}
\label{sec:fes}

\cite{fawkes:22} is an excellent paper about the challenges of causal modeling, particularly in the context of algorithmic fairness. Its emphasis is on selection bias, and I recommend it as a reading. I fully agree that selection bias is an extremely important aspect of modeling, and some of its challenges were mentioned already in our early review \cite{loftus:18}. If anything, \cite{fawkes:22} does not go far enough, not mentioning general problems of selection bias in behavioral modeling by machine learning: distribution shift, a major problem for predictive modeling regardless of any causal or fairness issues, can in part be attributed to frequent changes in selection bias from training to test time.

However, one of the main barriers for addressing these problems in \cite{fa-fast:00} appears to be an incorrect belief that counterfactual fairness \emph{requires} ancestral closure of protected attributes: the notion that if some variable $A_i$ is a protected attribute, then any causal ancestor of $A_i$ must also be by default. This sometimes make sense, sometimes does not, but \cite{fawkes:22} go as far as claiming that ``In the case of counterfactual fairness ancestral closure is mentioned as an explicit requirement on the set of sensitive attributes.'' However, ancestral closure of protected attributes was never a fundamental property of counterfactual fairness.

\paragraph{Ancestral closure of protected attributes was never a fundamental property of counterfactual fairness.} Let us start with a verbatim quote from \cite{kusner:17}, page 5:

\begin{quote}
    Suppose that a parent of a member of $A$ is not in $A$. Counterfactual fairness allows for the use of it in the definition of $\hat Y$.
\end{quote}

As spelled out above, in no ambiguous terms, there isn't at all any default requirement for ancestral closure. However, continuing from the same paragraph (emphasis added),
\begin{quote}
\emph{If} this seems counterintuitive,
then we argue that the fault should be at the postulated set of protected attributes rather than with the definition of counterfactual fairness, and that typically we should expect set $A$ to be closed under ancestral relationships given by the causal graph. 
\end{quote}

What if it does \emph{not} seem counterintuitive to keep an ancestor of $A$ out of the set of protected attributes? \emph{Then we just keep it out.} A modeler can augment their initial choice of $A$ after seeing the causal model (as the causal model can be built independently without judgments of which attributes are protected): it could be some ancestors of the initial $A$, some non-ancestors, no ancestors etc. However, counterfactual fairness does \emph{not} dictate what counts as a protected attribute: that's a value judgement not be left to any automated procedure. $A$ is taken as a primitive, an input to the definition and algorithms.

The paragraph about ancestral closure of protected attributes came about after a discussion with reviewers from ICML 2017, to where the paper was originally submitted. This was one way of acknowledging their input, which helped to improve the clarity of the manuscript, but I take the responsibility for introducing this paragraph in a confusing way. The reason why the authors of \cite{fawkes:22} and others (such as \cite{plecko:22}) got to conclude that ancestral closure is fundamental may be due to the formatting of Section 3.2 of \cite{kusner:17}, titled ``Implications'': the title of the respective paragraph (``Ancestral closure of protected attributes'') should have been ``Counterfactual fairness allows for the use of ancestors of $A$ in the definition of $\hat Y$''. The whole closure concept was meant only as an \emph{example} of why one may want to revisit their choice of $A$ by thinking causally! Surely it makes no sense to claim that Definition 5 in \cite{kusner:17} implies ancestral closure.

Finally, it is irrelevant whether \cite{fawkes:22} and \cite{rosenblatt:23} found that ``(ignorability) is also assumed in most DAGs we can find in the literature''. The papers cited in \cite{fawkes:22} were introducing novel concepts where examples are kept simple to focus the attention of the reader. Selection bias, lack of ignorability and partial identifiability are all issues that are core to causal modeling. Counterfactual fairness added a link between causal modeling and algorithmic fairness, always fully acknowledging that it inherits all the demands that come with an appropriate account of causality.

\paragraph{What is counterfactual fairness for, after all?} \cite{fawkes:22} contains a very interesting example that is worth discussing (emphasis added):

\begin{quote}
    We can imagine scenarios when treating counterfactuals like this could be intuitively very unfair. As an example, again we look at law school admissions, but now our sensitive attribute $A$ is the presence of a particular disability with severe adverse effects; for instance, it might mean that, on average, sufferers can only work or study for half the amount of time per day than someone who does not have this disability. If an individual were able to attend college to get a GPA, take the LSAT, and perform well enough in both of these to apply to law school despite having this disability, it is reasonable to assume that they are an exceptional candidate and would have performed exceptionally well relative to all candidates had they been born without any disability. \emph{However, their structural counterfactuals formed as above would look like an average applicant born without the disability. Therefore a predictor which is counterfactually fair relative to these structural counterfactuals would simply treat this candidate as an average applicant without the disability.} This does not capture an intuitive notion of fairness in this scenario. Further it does not align with what we imagine counterfactual fairness as doing. The structural counterfactuals are failing to correcting for the difficulties of having this sensitive attribute, which is one of the main appeals and claims of counterfactual fairness.

\end{quote}

I am puzzled by the above. Using a counterfactually fair predictor is one way of \emph{not} to ``simply treat this candidate as an average applicant without the disability.'' 

To see this, let $X_a$ be the (say) GPA score had the candidate have the disability and let $X_{a'}$ be the corresponding score had the candidate not have the disability. One way of constructing a counterfactually fair predictor is to start from a postulated structural equation $X = f(A, U)$ and build the predictor around $U$. To make my argument more explicit, I will build $\hat Y$ using only $f(a, U)$ and $f(a', U)$ a.k.a. $X_a$ and $X_{a'}$ -- this has the advantage of avoiding the worry about the dimensionality of $U$, for instance, which may as well be infinite-dimensional, as well as avoid speculating about the meaning of $U$, like ``grit index''\footnote{While we favored interpretations of latent variables in \cite{kusner:17}, and they can be useful in Level 2 modeling where an explicit measurement model is provided, they are in general for didactic motivation only.}. For most reasonable outcomes $Y$, it is expected that $\hat Y$ is a non-decreasing monotone function of both $X_a$ and $X_{a'}$ (say, $\hat Y := \lambda_1X_a + \lambda_2X_{a'}$ for positive coefficients $\lambda_1$ and $\lambda_2$, as an example). 

So say we have two candidates, 1 and 2, who obtained exactly the same score $X^{(1)} = X^{(2)} = x$, where one comes from the stratum $A = a$ ($A^{(1)} = a$) and the other from $A = a'$ ($A^{(2)} = a')$. Under the assumption informally made by \cite{fawkes:22}, $X^{(i)}_{a'} \geq X^{(i)}_a$ for any individual indexed as $i$, an assumption that can be hard-coded in the counterfactual model (monotonicity assumptions have a long history in cross-world causal analysis). 

As $X^{(1)}_{a'} \geq X^{(1)}_a = x = X^{(2)}_{a'} \geq X^{(2)}_a$, it follows that $\hat Y^{(1)} \geq \hat Y^{(2)}$. The exceptionalism of candidate 1 has been recognized by counterfactual fairness as a straightforward consequence of the definition and the causal model, contrary to what could be interpreted from the quote above.

But let me use this example to discuss a more fundamental (and better motivated) criticism of counterfactual fairness that goes beyond technical questions of causal assumptions. A party with stakes at the college admission problem may, sensibly, wonder whether we are doing the candidate from group $A = a$ any favors, if the \emph{real} outcome $Y$ of interest e.g. successful graduation, still has lower odds of happening compared to incoming students from group $A = a'$. Our prediction may lump together $X_a$ and $X_{a'}$ for all candidates when building $\hat Y$, but it still does not change the fact that life continues to be hard once the candidate becomes a student, and challenges for outcome $Y$ (again, not $\hat Y$) are of the same nature as challenges for pre-requisite $X$. Perhaps such candidates of such characteristics may be better served by a less demanding (even if less prestigious) alternative program with higher chances of success.

As a another example, suppose that as part of a loan decision making pipeline we want the prediction of whether a business case will default the loan or not. One factor may be the ethnic background of the business owner, who may suffer from prejudice in the community where the business will take place. Suppose we can assume as a fact that odds are stacked against this individual. A counterfactual fair predictor will not single out just the observed characteristics of the applicant, but the whole set of counterfactual scenarios regardless of which one is the real one. It is still the case, though, that the world is unfair and odds are compromised. Are we doing this applicant any favor, given that the risk is real?

One idea is to hard-code a compromise, accept a candidate from a disadvantage background if their score is ``just so'' below the threshold of privileged groups. But setting such a fudge-factor threshold is not trivial and may raise contention, while at least in principle counterfactual fairness provides a constructive path to how predictions should take place. And regardless of the decision method, it is \emph{still} a fact they are at a disadvantage, with all the consequences that are entailed.

Counterfactual fairness plays to the concept that we should not be \emph{fatalist} about the outcome. Yes, the business owner has a higher chance of failing, but they are not predestined to fail. A reality check can still be done with the factual-only prediction about extreme cases that should be filtered out. Counterfactual fairness scores can still be used to rank candidates that pass this triage into the next stage of the pipeline. 

But, more importantly, perhaps society and institutions should be better prepared to \emph{share the burden of failure}. It is convenient, and understandable, for a bank to follow the fatalist route and refuse a loan to those cases which do not reach the threshold, with or without a fudge factor to account for less privileged cases, as the applicant may otherwise end up struggling with debt and miss out on alternatives to the possibly-doomed business plan. An admissions officer in a competitive university program may have learned from experience that candidates from some underprivileged group may struggle and fail in their studies, wasting time and money in their lives. The officer may conclude that the moral thing to do is to statistically protect this individual by denying acceptance unless they show exceptional evidence that they are especially prepared. However, the above is predicated on the burden of failure mostly falling on the shoulders of the applicant: it is their time to lose, their money at stake. An alternative is being better prepared to \emph{spread the risk}: the provision of government or charity funds, or even tax incentives to banks, to mitigate the stakes put forward by a loan applicant from an underprivileged background who is partially set up to fail for issues out of their control; fast-tracking schemes for underprivileged candidates in a demanding university program to transfer to another one that better accommodates their personal situation when needed. None of this is an easy path to follow, but the first and humble step is to acknowledge that merely predicting a biased-by-construction outcome $Y$ ``as is'' can end up being a perversion of how machine learning algorithms are meant to serve the public. The main fights are on the non-technical aspects of decision-making which are societal, not technological, considerations, but this does not mean getting a free pass to ignore the causal effects of someone's membership in an underprivileged group against the criteria by which we constantly judge such an individual.

\section{Comments on Plecko and Bareinboim (2022)}

The survey by Plecko and Bareinboim \cite{plecko:22} is a terrific resource for those interested in the uses of causal modeling in algorithmic fairness. I strongly recommend it! Among its broad scope, Section 4.4.1 in that paper covers counterfactual fairness. Unfortunately, that section doesn't quite get right some its key aspects. In what follows, I will touch upon a sample of the main issues.

\paragraph{Pearl's ``notational confusion''.} Section 4.4.1 of \cite{plecko:22} starts by challenging our notation in \cite{kusner:17}, but the arguments don't add up. I do not believe there is a justifiable basis for the misunderstanding, as I will explain soon. I do believe, though, that the difficulties that led to the misunderstanding has origins on Pearl's unorthodox way of denoting counterfactuals, so let's start with it.

The classic paper by Holland \cite{holland:86} popularized the notation ``$Y_a(u)$'' to denote the potential outcome of variable $Y$ for \emph{unit} $u$ under intervention $A = a$. In Holland's notation, units come from an population $U$, not to be confused by the usual symbol used to denote background variables in the structural causal model (SCM) framework. In Holland's notation, \emph{all that it means is an abstract way of indexing data points.} For instance, using $V^{(i)}$ to denote the $i$-th sample of some variable $V$, we can denote a dataset of three data points as $$\{(A^{(1)}, X^{(1)}, Y^{(1)}), (A^{(2)}, X^{(2)}, Y^{(2)}), (A^{(3)}, X^{(3)}, Y^{(3)})\}.$$

If this dataset is our population, we can think of Holland's $u$ as nothing more than a placeholder for 1, 2, 3, and $U$ as nothing more than $\{1, 2, 3\}$. This is particularly suitable for the type of analysis that researchers like Rubin, Holland, Rosenbaum and several others have been interested on across the years, following Fisher's original framework for randomized controlled trials (RCTs): there is (typically) a finite population $U$ (say, the people in the RCT), the potential outcomes are fixed, \emph{and all the randomness is in the treatment assignment}. By construction, this is a framework with zero out-of-sample predictive abilities\footnote{Think of the analogue in survey sampling: instead of, say, forecasting who would win an election, consider the problem of  estimating a snapshot of how many votes a candidate would get today assuming fixed vote intentions, where the role of randomization is to decide who gets to be asked about their choice on the day. In classic Fisherian RCT analysis, ``who gets to be asked'' corresponds to which potential outcome gets to be probed.} ``Snapshot analysis'' is of course useful for some problems of algorithmic fairness such as  auditing a fixed dataset/finite population, but clearly not the point of \emph{predictive} algorithmic fairness for out-of-sample datapoints.

When Pearl introduced his framework \cite{pearl:09}, he overloaded the meaning of $U$ in his notation: both as background latent variables with a probability distribution, \emph{and} as an abstract data point index! This can easily confuse any reader, even more so those used to Rubin/Holland's notation. For an example, consider Chapter 4, p. 92, of Primer \cite[Pearl's website edition]{pearl:16}, 
\begin{quote}
It is in this sense that every assignment $U = u$ corresponds to a single member or ``unit'' in a population, or to a ``situation'' in ``nature''.    
\end{quote}
In that chapter, $U$ bounces back from being a random variable (Eq. 4.3 and 4.4), to being a Hollandesque data point index (``... corresponds to a single member...''), to being a realization of a random variable (what else ``$U = u$'' is trying to convey if not the realization of a random variable, using the standard notation of all probability textbooks?). It is all very confusing\footnote{Not confusing enough? Consider the fact there is no logical reason why two units cannot have the same value for $U$, i.e., units $i$ and $j$ could very well have realizations $U^{(i)} = U^{(j)} = u$. Sure, if we assume $U$ is continuous, this can only happen with probability zero. But measure-theoretical cop-outs should not stop us from pointing out the logical shortcomings of statements such as ``...every assignment $U = u$ corresponds to a single member or ``unit'' in a population...''. } -- but notice that I wrote the title of this section in scare quotes (``notational confusion'') because I'm sure Pearl understands the differences, and all he tries to do is a type of notational overloading that is sometimes helpful\footnote{That is not even Pearl's most famous unusual twist of notation: a statement such that $p(y~|~do(x), z)$ is itself a source of confusion for the more orthodox-minded statistician, as it suggests that ``do(x)'' is a realization of a random variable as opposed to the index of a regime. The orthodox notation would be something along the lines of $p_{do(x)}(y~|~z)$ or $p(y~|~z; do(x))$, but it is easy -- at least for me -- to appreciate Pearl's choice of notation, as it leads to more syntactically elegant manipulations when deriving identification results using the do-calculus.} in some of his derivations and definitions. For instance, as far as I know, he never writes $Y_x(U)$, with capital $U$, suggesting that $Y_x(u)$ is more of a Holland-style incarnation of $u$ as an ``unit'' index, as opposed to the realization $u$ of random variable/process $U$.

Why did I spend time in this detour? Because I think many people in causal machine learning do not appreciate it, partially because it is a minutia that does not mean much in practice. Nevertheless, \cite{plecko:22}, probably by trying to orthodoxally follow Pearl's unorthodox notation, gets our definition (5) from \cite{kusner:17} very wrong. We use $\hat Y_{A \leftarrow a}(U)$ to denote, constructively, that $\hat Y$ can (and usually should) be a function of background variables $U$ -- \emph{it is capital $U$, not lower case $u$} -- and in potential outcome notation it should very clearly be equivalent to 
\begin{equation}
P(\hat Y_a \leq y~|~A = a, X = x) = P(\hat Y_{a'} \leq y~|~A = a, X = x),
\label{eq:po_formulation}
\end{equation}
which I believe is what Eq. 198 of \cite{plecko:22} is meant to capture. If the use of capital letters is not clear enough to indicate that $\hat Y_{A \leftarrow a}(U)$ is a random variable, then the explicit sampling of $U$ in the {\sc FairLearning} algorithm should send the point home. Unlike my writing about ``ancestral closure'' (see previous section) which could induce confusion, I do not think this confusion in \cite{plecko:22} is justifiable.

As a matter of fact, prior to the first draft of \cite{kusner:17}, I considered whether to go with formulation (\ref{eq:po_formulation}) instead of the one with explicit background variables $U$. In the end, I suggested the one found in the final paper only because it more explicitly and constructively linked to algorithms in which the abduction step of Pearl's SCM framework is implemented. But it is (conceptually) immaterial whether one chooses to use $U$ or the whole principal stratum $\{X_a, X_{a'}, X_{a''}, \dots\}$ as inputs to $\hat Y$. We just did this in an example in the previous section.

\paragraph{Is counterfactual fairness an individual fairness notion? Yes.} Further confusion follows in \cite{plecko:22}, which challenges our claim of counterfactual fairness as an individual, as opposed to group, notion of fairness. Basically, the argument is that, because we don't adopt $\hat Y_a(u) = \hat Y_{a'}(u)$ as the definition, then our definition is not ``individual-level enough'' (for the lack of a better term).

As a matter of fact, we discuss all of these differences in Appendix S1 of \cite{kusner:17}, which is not mentioned by \cite{plecko:22}. Given that our Appendix S2 on demographic parity was summarily ignored by \cite{rosenblatt:23}, I can only conclude that reading appendices of NeurIPS papers is not a popular pastime. In the interest of being self-contained, I will paraphrase S1 here, noting that what follows below will be familiar those who read the original counterfactual fairness paper in its entirety.

Put simply, I thought that defining counterfactual fairness as $\hat Y_a(u) = \hat Y_{a'}(u)$ was a bad way of introducing a novel approach for looking into \emph{predictive} algorithmic fairness. The definition that was ultimately used in the paper came about because we wanted to make explicit that fairness through unawareness was a silly concept, as conditioning on $A$ directly is actually important; and we wanted to show that an alternative idea such as \cite{nabi:18} was overly conservative, as it forbade conditioning on mediators. What was a way of explicitly acknowledging that conditioning on $A$ and $X$ was not only allowed, but fundamentally a core characteristic of counterfactual fairness? That's right, Eq. (5) as originally introduced in \cite{kusner:17}, which can be interpreted as \emph{equality in distribution} between any pair $\hat Y_a$ and $\hat Y_{a'}$ under the measure induced by conditioning on $(A = a, X = x)$.

Is this still fundamentally an individual-level quantity? Of course it is. As described by Holland, Rubin and others, interventional contrasts obtained by conditioning on mediators, such as $$\mathbb E[\hat Y_a~|~x] - \mathbb E[\hat Y_{a'}~|~x] = \mathbb E[\hat Y~|~x, do(a)] - \mathbb E[\hat Y~|~x, do(a')]$$ are ``not even wrong'', in the sense that they do not define a ``causal effect'' as normally understood as \emph{comparing interventions at an individual level.} To be clear, a predictive distribution conditioned on a mediator, such as $\mathbb E[\hat Y~|~x, do(a)]$ \emph{is} well-defined: for instance, in a control problem/reinforcement learning scenario, this may tell us whether we should do a ``course correction'' towards a desirable outcome $\hat Y$ given that we controlled $A$ at level $a$ in the past and we just observed a recent state $X = x$. However, the \emph{contrast} $\mathbb E[\hat Y~|~x, do(a)] - \mathbb E[\hat Y~|~x, do(a')]$ is non-sensical, since a same person in general will not have $X = x$ under both scenarios $A = a$ and $A = a'$. At best, the contrast $\mathbb E[\hat Y~|~x, do(a)] - \mathbb E[\hat Y~|~x, do(a')]$ compares \emph{two different groups of people who happen to coincide at $X = x$}. Completely unlike that, the cross-world counterfactual estimand $\mathbb E[\hat Y_a~|~a, x] - \mathbb E[\hat Y_{a'}~|~a, x]$, among any other functionals of the respective distributions, is not only a well-defined but fundamentally an unit-level contrast, unlike what \cite{plecko:22} may imply.

Now, do we miss anything in the sense that we can have Eq. (\ref{eq:po_formulation}) while $Y_a(u) \neq Y_{a'}(u)$? Yes -- in essentially useless scenarios. Consider the case where $A \in \{0, 1\}$ and $\hat Y_0 := g(X_0, X_1, U_{dummy_0})$ and, for whatever reasons unbeknownst to you and me,  $\hat Y_1 := g(X_0, X_1, U_{dummy_1})$ where $U_{dummy_0}$ and $U_{dummy_1}$ are \emph{different} variables but \emph{equal in distribution} -- a construction not entirely dissimilar to the basic example in Section \ref{sec:background} used to debunk the equivalence between demographic parity and counterfactual fairness, except that here $U_{dummy_\star}$ is outside the structural model for $X$. Then $\hat Y_0$ and $\hat Y_1$ are equal in distribution, but (using now $u$ in the sense of ``unit'', just to illustrate Pearl's ``notational confusion'') $\hat Y_0(u) \neq \hat Y_1(u)$. My judgement is that sacrificing the explicit link to the observed values of $A$ and $X$ found naturally in the equality-in-distribution definition of counterfactual fairness, in order to tackle these cases, is an exercise in futility. 

I would also point out that our Appendix S1 goes beyond the variations described by \cite{plecko:22} and mentions the ``almost sure'' alternative definition of counterfactual fairness (that is, $P(Y_a = Y_{a'}~|~A = a, X = x) = 1$), arguably a more interesting case than $Y_a(``u'') = Y_{a'}(``u'')$ as again it makes the conditioning on $A$ and $X$ explicit). As in the ``sure'' case, we judged this variation to be pointless.

\paragraph{What is counterfactual fairness for, after all? (Round 2).} I applaud the thoroughness and breadth of examples in \cite{plecko:22}, which provides a great service to readers interested in the causal aspects of fairness. That been said, Example 14 in \cite{plecko:22} does not quite make sense. Let's revisit it.

The framing is that $A \in \{0, 1\}$ is the gender of a prospective employee, $X$ is a measure of performance and $Y$ is the salary to be paid. The causal model (in our notation) is given by

\begin{itemize}
    \item $A = U_a,$
    \item $X = -A + U_x,$
    \item $Y = A + X + U_y.$
\end{itemize}

\noindent where $U_a$ is uniformly distributed in $\{0, 1\}$ and error terms are mutually independent. With some basic algebra we can show that $Y_0$  and $Y_1$ are equal in distribution given $A$ and $X$. By interpreting this equality in distribution as counterfactual fairness, \cite{plecko:22} are dissatisfied, because:

\begin{itemize}
\item Gender 0 produces more than Gender 1 by equation $X = -A + U_x$. They ``should'' be paid more, but $Y$ is independent of $A$;
\item Seemingly outrageously, $Y$ further discriminates against genders, as $A$ shows up in the equation for $Y$.
\end{itemize}

There are quite a few conceptual problems in this analysis, at least to the extent they have anything to do with counterfactual fairness. To start with, framing it as counterfactual fairness is already ill-posed  as there is no prediction problem being specified: there is no $\hat Y$. A simulacrum of ``counterfactual fairness'' is used here to diagnose ``the fairness'' of an existing outcome variable that is set in place, a task which we did not aim at and in which we had little interest when we came up with the concept. This distinction matters, let's see why.

Let's ignore the $Y$ variable of the original example as the target of a fairness analysis -- the outcome can very well be biased, but we assume that there is nothing that we can do (at least immediately) about that. Instead, our starting point, if we were to \emph{choose} to build a total-effect counterfactually fair predictor\footnote{Just \emph{why} we would like to predict salaries is entirely lost on me. As I mentioned in previous sections (``let's predict college-level academic performance for school admissions, not placement offers''), the natural thing to do would be a prediction of performance, with salary offers following from a transparent deterministic rule of that prediction and other factors such as competitiveness. But let's go with it for the sake of keeping the same example.} of $Y$, is to define some $\hat Y := g(U_x)$ that best fits $Y$ given $(A, X)$ by, say, mean squared error. We get $\hat Y = U_x$, implying that $\hat Y$, given $A$ and $X$, is equal to $A + X$. It makes little sense to say that $\hat Y$ is automatically discriminatory by the virtue of being expressible as a function of $A$: the Red Car example that opens up our paper makes clear that in general we \emph{should} be using the protected attributes in the construction of the predictor in an indirect way, via a dependency on $U$ that can be re-expressed as a dependency on $A$ by conditioning.

If we are not happy to base our predictions on $U_x$ instead of $X$, we have nobody to blame but ourselves, since we \emph{chose} to hide the information about $A$ contained in $X$. The discussion in \cite{plecko:22} then goes on a tangent about the perils of total effect analysis, rigidly assuming that counterfactual fairness is bound to total effects by some Law of Nature, as opposed to our clear statement that \cite[page 5]{kusner:17}
\begin{quote}
...it is desirable to define path-specific variations of counterfactual fairness that allow for the inclusion of some descendants of $A$, as discussed by [21, 27] and the Supplementary Material\footnote{Who reads that, anyway? Jokes aside, the appendix on path-specific effects was indeed meant to be only a very light acknowledgement that a thorough coverage of the topic was yet to follow.}.
\end{quote}

The problem here has just one path: $A \rightarrow X$ (in counterfactual fairness as originally used, it doesn't make much sense to consider the paths into $Y$ -- to be clear, one \emph{can} use the same counterfactual fairness constraint with respect to any variable $Y$ of interest, but whether to judge the presence of $A$ in the structural equation for $Y$ as problematic or not will require further problem-specific justifications). It is not clear to me what exactly the path-specific problem we would like to get out of Example 14 is. I will assume (hopefully correctly) that the claim is that $A$ is protected, but path $A \rightarrow X$ is not. All this means is that $\hat Y$ can be a function of $X$, but not of $A$, prior to conditioning on $(A, X)$. This implies $\hat Y \indep A~|~(A, X)$ anyway. To the best of my educated guess, this recovers the intention of Example 14. 

To illustrate the conceptual simplicity of path-specific variants, let's also consider the following complementary problem: what if we had a graph $\{A \rightarrow X \rightarrow Z; A \rightarrow Z\}$ and we wanted the path $A \rightarrow X \rightarrow Z$ to be allowed to carry information from $A$, but not $A \rightarrow Z$? The game is the same: $X$ is free-for-all, but edge $A \rightarrow Z$ is broken, its tail replaced by fixed indices corresponding to any subset of the sample space of $A$ that we want. For instance, if $X = f_X(A, U_x)$, $Z = f_Z(A, X, U_z)$ are the structural equations, we can define $\hat Y$ as a function of $f_X(A, U_x)$ (a.k.a. $X$) and (any fixed subset of) $\{f_Z(a, X, U_z), f_Z(a', X, U_z), f_Z(a'', X, U_z)...\}$ (a.k.a. $\{Z_{aX}, Z_{a'X}, Z_{a''X}, \dots\}$, $X$ in the subscript indicating that $X$ is set to the value it would naturally occur anyway, so $Z_{aX} = Z_a$ etc.). Any resemblance to the idea of overriding protected attributes in (say) job applications, while propagating changes only to a subset of the content of the application, is not a coincidence. For the record, please notice that while the path-specific conceptualization may follow naturally, I would never say that implementing it is easy. Partial identification is arguably the way to go here even if challenging, with \cite{wu:19} being an early example. 

A shortcoming in the counterfactual fairness coverage of \cite{plecko:22} was the apparent belief that tackling the problem of \emph{diagnosing} unfairness was enough in order to comment on the problem of \emph{prediction}. It is not. To repeat myself, \emph{the primary goal of counterfactual fairness, as originally defined, is to provide a way of selecting information within $X$ that is deemed to be fair}, blocking ``channels'' of information from $A$ \emph{into $X$ only}, not into $Y$. We can't change $Y$ in the problem setup of \cite{kusner:17} (at least not in the short term, see \cite{loftus:18}), but we can change $\hat Y$. The very idea of using causal concepts in fairness problems is that is not enough to know \emph{which} values the measurements took, but also \emph{how} they came to be. By failing to acknowledging some of the ``how'' aspects of the target variable $\hat Y$ in counterfactual fairness, \cite{plecko:22} ends up not being sufficiently causal in its analysis.

\section{Conclusion}

There are connections between counterfactual fairness and demographic parity. We know that since the original paper \cite[Appendix S2]{kusner:17}. Further connections claimed by \cite{rosenblatt:23}, however, do not hold up. Counterfactual fairness does not imply demographic parity and demographic parity does not imply counterfactual fairness, not even approximately. It would be terrific if possible, but we cannot magically reduce definitions relying on Pearl's Rung 3 assumptions to Rung 1 or even Rung 2 assumptions. With the knowledge in this note, I hope that the interested reader can critically read papers such as e.g. \cite{anthis:23} and assess which gaps they may or may not have, and how to fill them up where needed with the necessary explicit assumptions.

I took the chance of highlighting other aspects of counterfactual fairness that are sometimes misunderstood. To summarize:
\begin{itemize}
    \item counterfactual fairness can be interpreted as a ``variable selection'' (information bottleneck/filtering) protocol for building predictors, agnostic to loss functions, other constraints and other decision steps. It arose to answer information selection questions that methods such as fairness through unawareness could not satisfactorily address. There is no reason not to build optimal predictors with respect to the information filtered by counterfactual fairness, subject to other constraints, if any. If one thinks counterfactual fairness by itself should suffice to magically solve arbitrary resource allocation requirements/Pareto optimalities/ranking constraints/other balances not already implied by the independence structure of the causal model, then they have seriously misunderstood it;
    \item ancestral closure of protected attributes was never a fundamental property of counterfactual fairness. I have no problems accepting that ambiguous writing in the original manuscript could lead to this confusion, but it is a matter of fact that ``counterfactual fairness allows for the use of (causal ancestors of protected attributes) in the definition of $\hat Y$'' \cite[page 5]{kusner:17}. 
\end{itemize}

\section*{Acknowledgements}

This work was partially funded by the EPSRC fellowship EP/W024330/1. I would like to thank Matt Kusner, Joshua Loftus and Chris Russell for helpful comments. Any mistakes are of my own making.

\bibliographystyle{plain}
\bibliography{rbas}

\appendix

\section*{Appendix: A note on the rank plot experiment in \cite{rosenblatt:23}.}

The experimental section in \cite{rosenblatt:23} argues that if we take, for instance, 40 samples from the subpopulation where $A = a$, then the rank statistics for $\bar Y$ and $\hat Y$ in the data are the same if we use Eq. (\ref{eq:real_deal}). This is not surprising as this mapping between $y^{(i)}$ and $\hat y^{(i)}$, a composition of invertible mappings, is rank-preserving. The experimental section then goes on  to point out that counterfactual fairness seems to return very different rankings in the particular study they present, which is seemingly a bad property -- but actually, as we will see, it is entirely expected. The plot in Figure 4 of \cite{rosenblatt:23} displays six methods for predicting first year average from race, sex and two types of exams in the law school dataset also used in \cite{kusner:17}. There are 40 columns of points representing a sample of 40 individuals recorded as \emph{Black} in the race attribute of the dataset. The columns are sorted according to the rank statistics of $\bar Y$ for the baseline method, a linear regression that includes all variables (``Full''). For each data point, each of six predictors is represented by a point, with points of a same predictor aligned in the same horizontal line. This makes the plot a $40 \times 6$ grid of points. For each pair of consecutive rows, a line segment connects points of two predictors that have the same marginal rank statistic.

Besides ``Full'', we have Listing 1 and Listing 2 predictors built by transforming the ``Full'' predictions. By construction, all of these three vectors of predictions have rank correlation of near 1 (it is not exactly 1 as e.g. ties in the predictor means that $\bar Y$, ``Full'', is not fully continuous). When plotted as the three bottom rows of Figure 4, this results in vertical line segments cutting across the three rows, as expected.

The top three rows, which are inspired by counterfactual fairness, are poorly aligned with the bottom three and among themselves. \emph{This is exactly what we should expect.} Each of the three ``counterfactually fair''-inspired methods apply a \emph{different} filter to the data: they use different subsets of information to build the predictor based on different assumptions\footnote{And just as a reminder, real-world-assumptions-free rules to decide which information in $X$ carries unfair consequences of $A$ amount to snake oil, in the view of this writer. Domain-independent buttons to be pushed are convenient for machine learning companies, less so for the people at the receiving ending of their services.}, and in a relatively low signal-to-noise ratio problem as the law school one. Poor rank correlation is what one should anticipate. It would have been an anomaly if in Figure 4 of \cite{rosenblatt:23} we saw strong agreements among predictors which use different subsets of information in this dataset.

To drive the point home, let's forget about counterfactual fairness and just redo the experiment with two choices of $\bar Y$: the true label $Y$, which is of course not usable in a real predictor and here will play the role of an oracle; and the full linear regression predictor as in \cite{rosenblatt:23}. This emulates two predictors which rely on correlated but different sources of information. We will call the corresponding ``fair'' algorithms ``Listing 2T'' and ``Listing 2F''. It is unclear why we would believe that alignment with an ``unfair'' predictor such as (say) full regression is of any relevance, if it shows such a poor relationship to the true ``fair'' alignment. Exemplar results are shown in Figure \ref{fig:experiments}. {\sc R} code to create plots of this type is shown below.

\begin{figure}[t!]
    \begin{tabular}{ccc}
       \includegraphics[width=5cm]{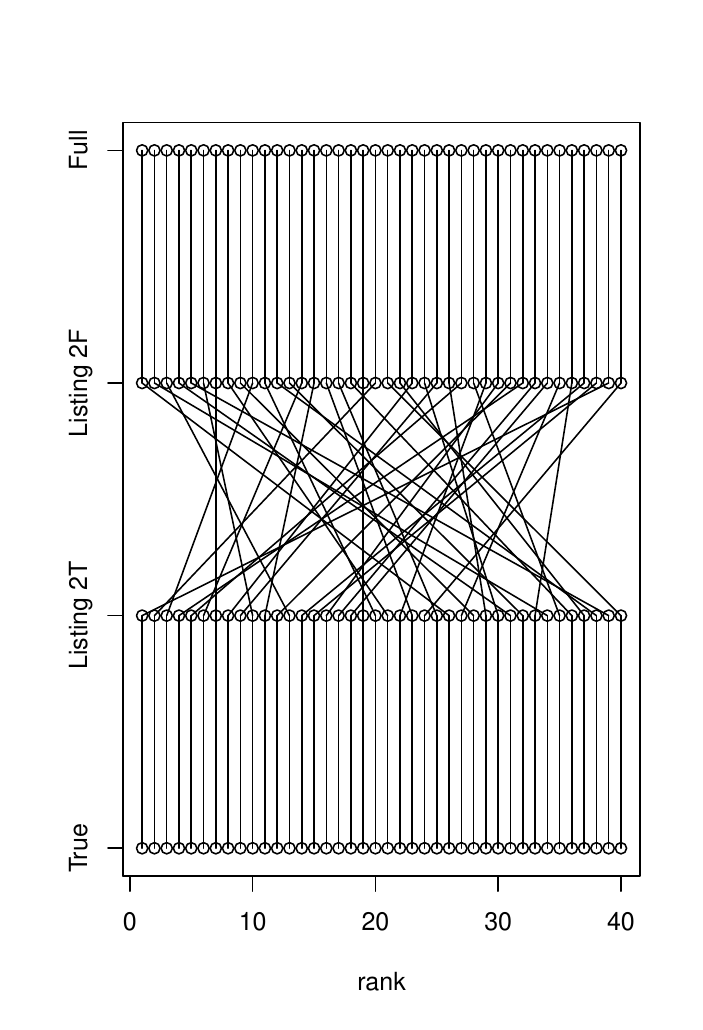} &
       \includegraphics[width=5cm]{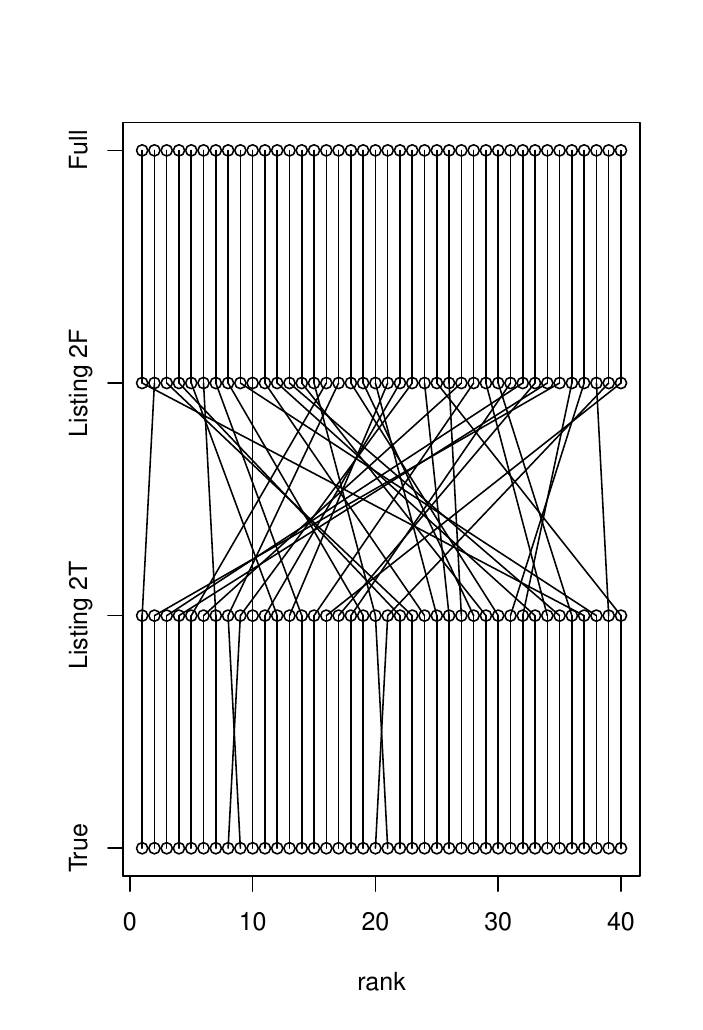} &
       \includegraphics[width=5cm]{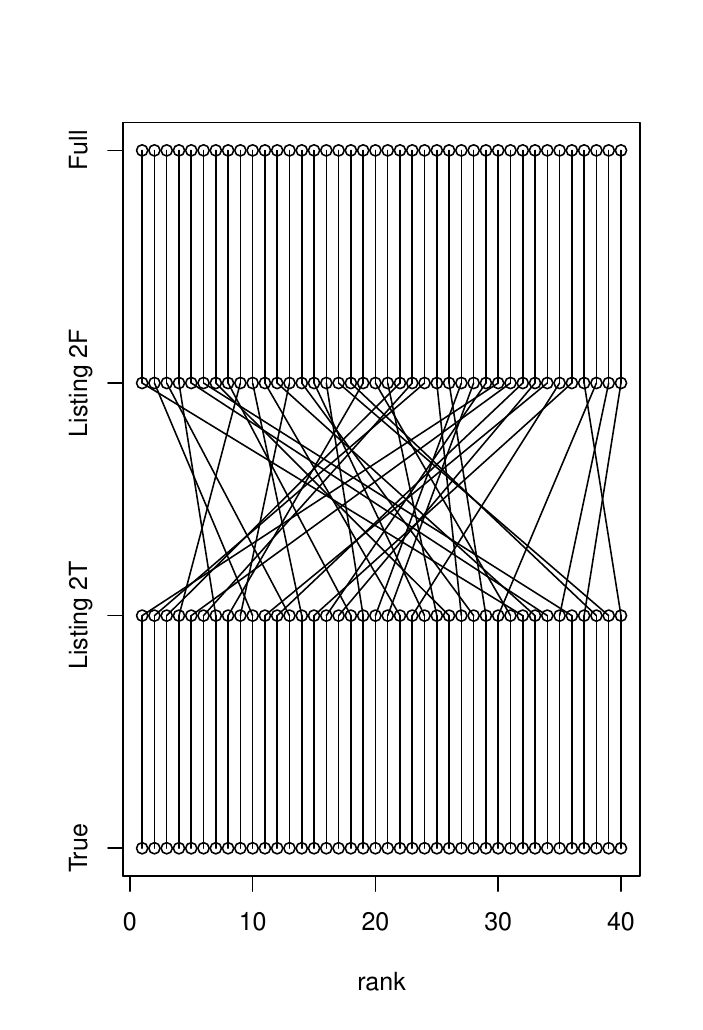}
     \end{tabular}
\caption{How stable is the rank of a (invertible function of a) predictor based on linear regression compared to the rank of the true outcome? Poor, as expected: the law school dataset has relatively low signal-to-noise ratio. Each plot above is generated using a different subset of 40 test points. There are 40 columns of points in each plot, sorted from left to right based on the value of the final row, ``True'', which is the corresponding true first year average. Between any two consecutive rows, we link points which have the same corresponding rank statistic as given by a dataset of four predictors sorted by the true first year average. Listing 2T is an invertible ``fair'' mapping of the ``unfair'' true value, and Listing 2F is an invertible ``fair'' mapping of the ``unfair'' linear regression prediction. The rank correlation between ``fair'' linear regression and ``fair'' true labels is relatively weak, casting doubt about the value added of shadowing any particular $\bar Y$. }
\label{fig:experiments}
\end{figure}

\begin{lstlisting}[language=R]

# Load the data and define train/test split. Data can be obtained from 
# https://github.com/mkusner/counterfactual-fairness
law_dat <- read.csv("law_data.csv")
subgroup_idx <- which(law_dat$race == "Black")

n <- nrow(law_dat)
train_idx <- sample(n, round(n / 2))
test_idx <- setdiff(1:n, train_idx)

# Fit model
lin_model <- lm(ZFYA ~ race + sex + LSAT + UGPA, data=law_dat[train_idx, ])

# Learn the cdfs of ZFYA stratified by race (just one subgroup in this case). 
# We also marginally sort ZFYA to be able to get empirical quantiles later on.
cond_black_cdf <- ecdf(predict(lin_model, 
                       law_dat[intersect(train_idx, subgroup_idx), ]))
cond_black_cdf_oracle <- ecdf(law_dat$ZFYA[intersect(train_idx, subgroup_idx)])
quantiles_marg <- sort(lin_model$fitted.values)
quantiles_marg_oracle <- sort(law_dat$ZFYA[train_idx])

# Select a test subsample of Black applicants, size 40
test_subsample_idx <- sample(intersect(test_idx, subgroup_idx), 40)

# Generate plain ("unfair") predictions
y_bar <- predict(lin_model, law_dat[test_subsample_idx, ])
y_bar_oracle <- law_dat$ZFYA[test_subsample_idx]

# Generate "fair" predictor based on the idea of ranking them as if they came
# from the aggregated distribution over races.
quantiles_hat <- round(length(train_idx) * cond_black_cdf(y_bar))
quantiles_hat[which(quantiles_hat == 0)] <- 1
y_hat <- quantiles_marg[quantiles_hat]
quantiles_oracle <- round(length(train_idx) * cond_black_cdf_oracle(y_bar_oracle))
quantiles_oracle[which(quantiles_oracle == 0)] <- 1
y_hat_oracle <- quantiles_marg_oracle[quantiles_oracle]

# Calculate rank correlations between possible labels and true label.
cat(paste("Spearman correlation ('unfair'): ", 
          cor(y_bar, y_bar_oracle, method="spearman")))
cat(paste("Spearman correlation ('fair'): ", 
          cor(y_hat, y_bar_oracle, method="spearman")))
cat(paste("Spearman correlation ('oracle'): ", 
          cor(y_hat_oracle, y_bar_oracle, method="spearman")))

# Plots.
#
# This works as follows. At the bottom we just plots the numbers 1-40 to 
# indicate that these are the possible ranks of the true label, as we have 40 
# demonstrative data points.
#
# In the second row from the bottom, we plot the ranks of each y_hat_oracle 
# corresponding to the ranks for the true label. By construction, this should be 
# the same, barring minor variability due to ties being broken arbitrarily.
#
# In the third row from the bottom, we plot the ranks of each y_hat 
# corresponding to post-processed full regression fits, as defined by the 
# Listing 2 algorithm. Since rank correlation is low here, this will obviously 
# be an entangled mess when linked to the points in the second row.
#
# The top row again has near perfect rank correlation with the row immediately
# below it, by construction.

rank_dat <- cbind(rank(y_bar_oracle, ties.method="first"), 
                  rank(y_hat_oracle, ties.method="first"), 
                  rank(y_hat, ties.method="first"),
                  rank(y_bar, ties.method="first"))
row_ordering <- sort(rank_dat[, 1], index.return=TRUE)$ix
rank_dat <- rank_dat[row_ordering, ]

# Define four sets of points
preds <- list()
for (i in 1:4)
     preds[[i]] <- data.frame(x = rank_dat[, i], y = rep(i, 40))

# Plot the points
plot(preds[[1]]$x, preds[[1]]$y, xlim = c(1, 40), ylim = c(1, 4), 
     xlab = "rank", ylab = "",  yaxt = "n")
axis(2, at = c(1, 2, 3, 4), 
     labels=c("True", "Listing 2T", "Listing 2F", "Full"))
for (i in 2:4)
     points(preds[[i]]$x, preds[[i]]$y)

# Draw lines between corresponding points in two consecutive vectors
for (n in 1:40)
    for (i in 1:3)
         lines(c(preds[[i]]$x[n], preds[[i + 1]]$x[n]), 
               c(preds[[i]]$y[n], preds[[i + 1]]$y[n]))

\end{lstlisting}

\end{document}